\pdfoutput=1
\documentclass{article}

\usepackage[letterpaper,top=2cm,bottom=2cm,left=3cm,right=3cm,marginparwidth=1.75cm]{geometry}

\usepackage[utf8]{inputenc} 
\usepackage[T1]{fontenc}    
\usepackage{hyperref}[]
\usepackage{url}            
\usepackage{booktabs}
\usepackage{natbib}
\usepackage{amsfonts}      
\usepackage{nicefrac}       
\usepackage{microtype}      
\usepackage{amsmath}
\usepackage{amssymb}
\usepackage[section]{placeins}
\usepackage{graphicx}
\usepackage{natbib}
\usepackage{booktabs}
\usepackage{colortbl} 
\usepackage{soul} 
\usepackage{tabularx}
\usepackage{hyperref}
\usepackage{stfloats}
\usepackage{algorithm,algorithmicx}
\usepackage[noend]{algpseudocode}
\usepackage{algcompatible}
\graphicspath{{./figures/}}
\usepackage{multirow}
\usepackage{mathtools}
\usepackage{shortvrb}
\usepackage{xspace}
\usepackage[most]{tcolorbox}
\usepackage[normalem]{ulem}
\useunder{\uline}{\ul}{}

\usepackage{amsmath,amsfonts,bm}

\def\eqref#1{equation~\ref{#1}}

\def\1{\bm{1}}

\def\rvx{{\mathbf{x}}}

\def\vtheta{{\bm{\theta}}}

\def\vx{{\bm{x}}}

\DeclareMathAlphabet{\mathsfit}{\encodingdefault}{\sfdefault}{m}{sl}
\SetMathAlphabet{\mathsfit}{bold}{\encodingdefault}{\sfdefault}{bx}{n}

\newcommand{\E}{\mathbb{E}}

\definecolor{lightgrey}{HTML}{dcdbdb}
\definecolor{lightblue}{HTML}{E8F0FE}
\definecolor{gray}{HTML}{9aa0a6}
\definecolor{lightpink}{HTML}{F48FB1}
\definecolor{lightred}{HTML}{FFCBC9}
\definecolor{lightcyan}{HTML}{80DEEA}
\newcommand{\cc}[0]{\cellcolor{lightblue}}
\newtcolorbox{mybox}[2][]
  {colback = black!5!white, colframe = black!75!black, fonttitle = \bfseries,
    colbacktitle = black!100!black, enhanced, 
    attach boxed title to top left={yshift=-2.2mm,xshift=4mm},
    title=#2,#1}

\newcommand{\ourmodel}{Dream\xspace}

\date{April 2, 2025}

\usepackage{soul}

\setcitestyle{round}

\newcommand{\github}{\raisebox{-1.5pt}{\includegraphics[height=1.05em]{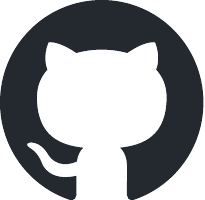}}\xspace}
\newcommand{\huggingface}{\raisebox{-1.5pt}{\includegraphics[height=1.05em]{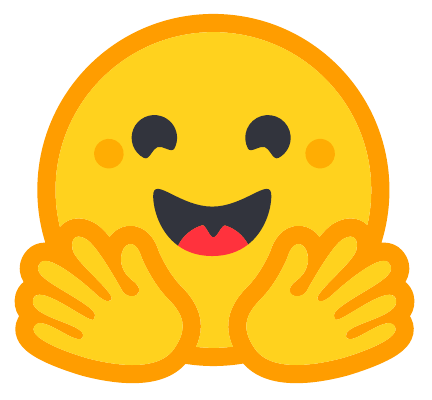}}\xspace}

\title{Dream 7B: Diffusion Large Language Models}

\author{\textbf{Jiacheng Ye}$^1$\thanks{Equal Contribution} \quad \textbf{Zhihui Xie}$^{1}$\footnotemark[1] \quad \textbf{Lin Zheng}$^1$\footnotemark[1] \quad
\textbf{Jiahui Gao}$^2$\footnotemark[1] \quad\textbf{Zirui Wu}$^1$ \quad \and
\textbf{Xin Jiang}$^2$ \quad \textbf{Zhenguo Li}$^2$ \quad
\textbf{Lingpeng Kong}$^1$\quad
\\
$^1$The University of Hong Kong\quad
$^2$ Huawei Noah's Ark Lab \\
  \github\href{https://github.com/DreamLM/Dream}{\textbf{DreamLM/Dream}} \quad \huggingface\href{https://huggingface.co/collections/Dream-org/dream-7b-68761d3d0665386f43f310c9}{\textbf{Dream-7B}}
}
  
\begin{document}

\maketitle

\begin{abstract}
We introduce Dream 7B, the most powerful open diffusion large language model to date. Unlike autoregressive (AR) models that generate tokens sequentially, Dream 7B employs discrete diffusion modeling to refine sequences in parallel through iterative denoising. Our model consistently outperforms existing diffusion language models on general, mathematical, and coding tasks. Dream 7B demonstrates superior planning abilities and inference flexibility, including arbitrary-order generation, infilling capabilities, and tunable quality-speed trade-offs.
These results are achieved through simple yet effective training techniques, including AR-based LLM initialization and context-adaptive token-level noise rescheduling.
We release both Dream-Base and Dream-Instruct to facilitate further research in diffusion-based language modeling.
\end{abstract}

\begin{figure}[h]
\centering
\includegraphics[width=1.0\textwidth]{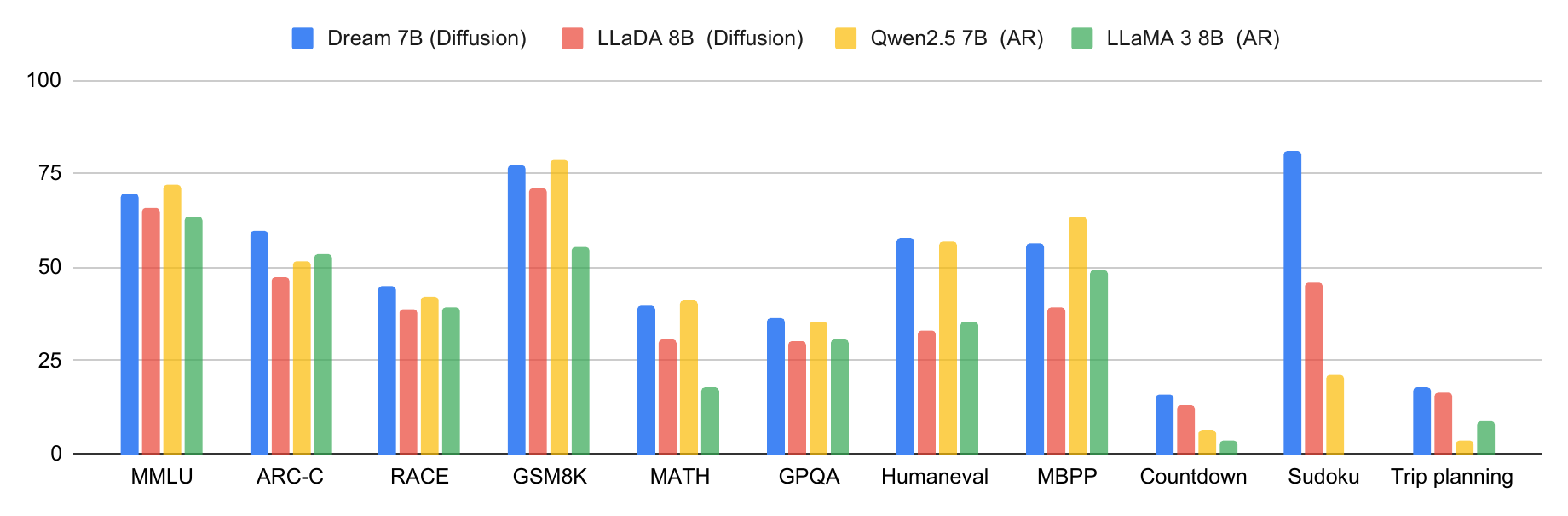}
\caption{
    \centering
     Performance comparison of diffusion and autoregressive language models on various tasks.
}
\label{fig:dream_performance}
\end{figure}

\section{Introduction}

The rapid advancement of large language models (LLMs) has fundamentally transformed artificial intelligence, driving breakthroughs across diverse applications and industries. The current landscape is dominated by autoregressive (AR) models that generate text through sequential left-to-right token prediction, with virtually all leading systems including GPT-4~\citep{openai2024gpt4technicalreport} and DeepSeek R1~\citep{guo2025deepseekR1}. 
While these models have demonstrated remarkable capabilities, a fundamental question emerges: What architectural paradigms might define the next generation of LLMs? This question becomes increasingly important as AR models exhibit inherent limitations in complex reasoning, long-term planning, and maintaining coherence across extended contexts~\citep{bubeck2023sparks,dziri2023faith,bachmann2024pitfalls,ye2024beyond}.
These limitations are particularly crucial for emerging applications such as embodied AI, autonomous agents, and long-horizon decision-making systems, where the ability to reason holistically about entire sequences rather than incrementally becomes paramount.

Discrete diffusion models (DMs) have gained attention as a promising alternative for sequence generation since their introduction to the text domain~\citep{austin2021structured,hoogeboom2021argmax,campbell2022continuous}. Rather than building text token by token from left to right, diffusion models begin with completely corrupted sequences and gradually denoise them through multiple refinement steps, enabling bidirectional context integration and more flexible generation patterns.
This architectural shift unlocks several theoretical advantages that directly address autoregressive limitations. The parallel nature of diffusion enables richer contextual modeling by incorporating information from all positions simultaneously, potentially leading to more coherent and globally consistent outputs~\citep{li2022diffusion}. Furthermore, the iterative refinement process naturally supports controllable generation~\citep{gulrajani2023likelihood,li2022diffusion} and provides flexible quality-speed trade-offs through adjustable inference steps, offering new dimensions for optimization for test-time scaling~\citep{snell2024scaling,muennighoff2025s1,geiping2025scaling} that complement existing techniques like long-thought chain-of-thought reasoning~\citep{jaech2024o1,guo2025deepseekR1}.

Recent developments have begun to demonstrate the practical viability of scaled diffusion language models. Works such as DiffuLLaMA~\citep{gong2025scalingdiffusionlanguagemodels}, LLaDA~\citep{nie2025llada} have successfully scaled discrete diffusion to billion-parameter regimes. While Mercury Coder~\citep{inceptionlabs_mercury}, as a commercial implementation, has demonstrated remarkable inference efficiency in code generation. This rapid progress, combined with the inherent architectural advantages of diffusion language modeling, positions these models as a promising direction for overcoming the fundamental limitations of autoregressive approaches. However, significant gaps remain in achieving performance parity with state-of-the-art autoregressive models like Qwen2.5~\citep{qwen2025qwen25technicalreport} across various general tasks~\citep{clark2018think,hendrycks2020measuring}.

In this work, we present Dream 7B, a 7-billion parameter diffusion language model that bridges this performance gap while demonstrating unique capabilities inherent to the diffusion paradigm. 
We introduce a comprehensive training framework that leverages AR-based LLM initialization and context-adaptive noise scheduling to scale diffusion language models. Through extensive evaluation, we establish that diffusion models can match AR-based LLM performance on general language tasks while providing distinct advantages in planning-intensive scenarios and offering unprecedented inference flexibility through arbitrary-order generation and dynamic quality-speed optimization.
We release both Dream-Base and Dream-Instruct to facilitate further research in this promising direction.

\begin{mybox}[colback=gray!10]{Takeaways}
    \begin{itemize}
        \item We release \ourmodel\ 7B, which consistently outperforms existing diffusion language models by substantial margins across diverse benchmarks.
        
        \item \ourmodel\ 7B achieves competitive performance with Qwen 2.5 on standard benchmarks (general language understanding, mathematical reasoning, and code generation) while exhibiting superior planning abilities and novel inference flexibility features that naturally emerge from the diffusion modeling paradigm.
    \end{itemize}
\end{mybox}

\section{Related Work}
\subsection{Diffusion Models for Text}
Diffusion models were originally developed for continuous domains like image generation~\citep{song2020denoising,ho2020denoising}.
Building on their remarkable success, a number of works have extended diffusion techniques to text generation tasks by modeling text in a continuous embedding space, building continuous diffusion language models~\citep{li2022diffusion, gong2022diffuseq, gong-etal-2023-diffuseq, dieleman2022continuous}. 
Among these approaches, \citet{genie2023} demonstrated the feasibility of applying diffusion models within a pre-training and fine-tuning framework on a small-scale transformer.

To fit the discrete nature of text, \citet{austin2021structured, hoogeboom2021argmax, campbell2022continuous} introduced discrete diffusion, which operates on the discrete vocabulary space, establishing the foundation for discrete diffusion methods.
In this formulation, diffusion forward steps progressively with either \texttt{[MASK]} tokens (leading to an absorbing state) or random tokens (resulting in a uniform state), and the reverse process reconstructs the original text from these noised sequences. Building on this foundation, several studies have improved the training objectives for discrete diffusion~\citep{Zheng2023ARD,lou2023discrete, shi2025simplifiedgeneralizedmaskeddiffusion, ou2024your, zhao2024improving, sahoo2024simple}. 
Besides, \citet{he-etal-2023-diffusionbert, ye2025diffusionlanguagemodelsperform} have shown benefits by initializing discrete diffusion models with pre-trained masked language models (MLMs, e.g., BERT; \citealt{devlin-etal-2019-bert}).
Recent studies have highlighted diffusion language models' advantages for complex reasoning and planning tasks~\citep{ye2024diffusion,ye2024beyond, ye2025implicit}. Moreover, combining diffusion and autoregressive methods through block generation has emerged as a promising approach, demonstrating a flexible trade-off between parallelism and sequential coherence~\citep{ye2024diffusion,arriola2025blockdiffusion}.

\subsection{Diffusion Large Language Models}
\citet{gulrajani2023likelihoodbased} analyzed the scaling laws for continuous diffusion language models, highlighting that continuous diffusion models require substantially longer training compared to autoregressive (AR) counterparts for optimal computational efficiency. In the discrete domain, \citet{lou2023discrete} showed that masked diffusion models (MDMs) could achieve perplexities comparable to or exceeding GPT-2-level AR models.
Encouraged by these results, recent efforts have scaled diffusion language models to billions of parameters.
\citet{nie2025scalingmaskeddiffusionmodels} demonstrated the applicability of 1.1B-scale MDMs in language tasks such as question answering.  Instead of training from scratch, \citet{gong2025scalingdiffusionlanguagemodels} proposed adapting existing AR language models (GPT-2 and LLaMA, from 127M to 7B parameters) into diffusion-based counterparts termed DiffuGPT and DiffuLLaMA. Concurrently, \citet{nie2025llada} presented LLaDA, an 8B-parameter diffusion language model trained from scratch, which achieved performance competitive with LLaMA3-8B~\citep{grattafiori2024llama3herdmodels}. 
Furthermore, Mercury Coder~\citep{inceptionlabs_mercury} has recently shown commercial applicability and efficiency in code generation tasks. Overall, our work builds upon the masked diffusion paradigm and introduces new training techniques to push the performance further, highlighting the potential of diffusion-based models as an alternative to autoregressive LLMs.

\section{Preliminary}

\label{sec:bg}
\subsection{Auto-regressive Modeling}
Consider a sequence $\rvx \coloneq (\vx^1,\dots,\vx^N)$ of length $N$ sampled from a data distribution $q(\rvx)$. A fundamental approach in sequence modeling decomposes the joint probability of tokens into a product of conditional probabilities~\citep{jelinek1980interpolated,bengio2000neural}:
\begin{equation}
p_\vtheta(\rvx)=p_\vtheta(\vx^1)\hspace{-5mm}\underbrace{\prod_{n=2}^{N}p_\vtheta(\vx^n \mid \vx^{1:n-1}),}_{\text{progressive left-context prediction}}
\label{eq:prob_ar}
\end{equation}
where $\vtheta$ represents the model parameters and $\vx^{1:n-1}\coloneqq \vx^1, \dots, \vx^{n-1}$ denotes the preceding tokens.

\subsection{Discrete Diffusion Modeling}

Discrete diffusion models~\citep{sohl2015deep,hoogeboom2021argmax,austin2021structured} are a class of latent variable models characterized by a forward noising process and a learned reverse denoising process. Let $\rvx_t$ denote the noised sequence at timestep $t$ and the \texttt{[MASK]} noise is employed. The forward process $q(\rvx_{1:T}|\rvx_0) = \prod_{t=1}^T q(\rvx_t|\rvx_{t-1})$ progressively corrupts the original data $\rvx_0 \coloneqq \rvx$ into a sequence of increasingly noisy tokens (replaced with \texttt{[MASK]} tokens) $\rvx_{1:T}\coloneqq \rvx_1, \dots, \rvx_T$. The forward process continues until the sequence is fully denoised at $t = T$. For $t \in (0, T)$, the sequence $\rvx_t$ is partially noised using a noise schedule $\alpha_t$ (where for each token the probability of remaining unmasked is $\alpha_t$).
The backward process learns to gradually denoise the masked sequence back to the original data distribution by iteratively predicting masked tokens as $t$ moves from $T$ to $0$. This denoising process can be modeled as:
\begin{equation}
\label{eq:prob_dm}
    p_{\vtheta}(\rvx)=\sum_{\rvx_{1:T}\sim q}p(\rvx_T)\hspace{-8mm}\underbrace{\prod_{t=1}^Tp_{\vtheta}(\rvx_{t-1}|\rvx_t).}_{\text{progressive full-context prediction}}
\end{equation}
The model parameters $\vtheta$ can be optimized by minimizing the negative log-likelihood of the clean data $\rvx_0$. 

The discrete time formulation $t \in (0, T)$ restricts $\rvx_t$ to predetermined noise levels, which can introduce bias. To mitigate this limitation and enable flexible sampling across arbitrary noise levels, we adopt continuous-time parameterization where $t$ ranges continuously over $[0, 1]$~\citep{kingma2021variational,campbell2022continuous,shi2025simplifiedgeneralizedmaskeddiffusion,ou2024your}. This continuous formulation allows the forward process to be expressed as $q(\rvx_t|\rvx_s)$ for any $0 \leq s < t \leq 1$.  We employ discrete diffusion with absorbing states and simplify the objective as weighted cross-entropy losses: 
\begin{equation}
    L(\theta) = -\E_{\rvx_0 \sim q(\rvx),t\sim \mathcal{U}(0,1), \rvx_t\sim q(\rvx_t|\rvx_0)} w(t) \sum_{n=1}^N \1_{[\mathrm{\rvx}_t^n=\texttt{MASK}]} \log p_{\theta}(\rvx_0^n|\rvx_t),
\label{eq:loss_dm}
\end{equation}
where the indicator function $\1_{[\mathrm{\rvx_t^n}=\texttt{MASK}]}$ ensures that the loss is computed only on masked token positions, and 
$w(t) \in (0,1]$
is a time-dependent reweighting term determined by a transformation of $\alpha_t$. 
For instance, when the forward process adopts the noise schedule $\alpha_t = 1-t$, the corresponding weight becomes $w(t) = \frac{1}{t}$~\citep{austin2021structured,Zheng2023ARD},  which assigns greater importance to denoising steps closer to the clean data (i.e., as $t \to 0$). 
As demonstrated in~\citet{shi2025simplifiedgeneralizedmaskeddiffusion,gong2025scalingdiffusionlanguagemodels}, this weighted cross-entropy objective Equation.~(\ref{eq:loss_dm}) represents an effective reformulation of the ELBO, which enables optimization of the mask predictor $p_\vtheta(\cdot|\rvx_t)$ by providing a tractable variational upper bound on the negative log-likelihood.

\section{Approach}
\begin{figure}[h]
    \centering
    \includegraphics[width=\textwidth]{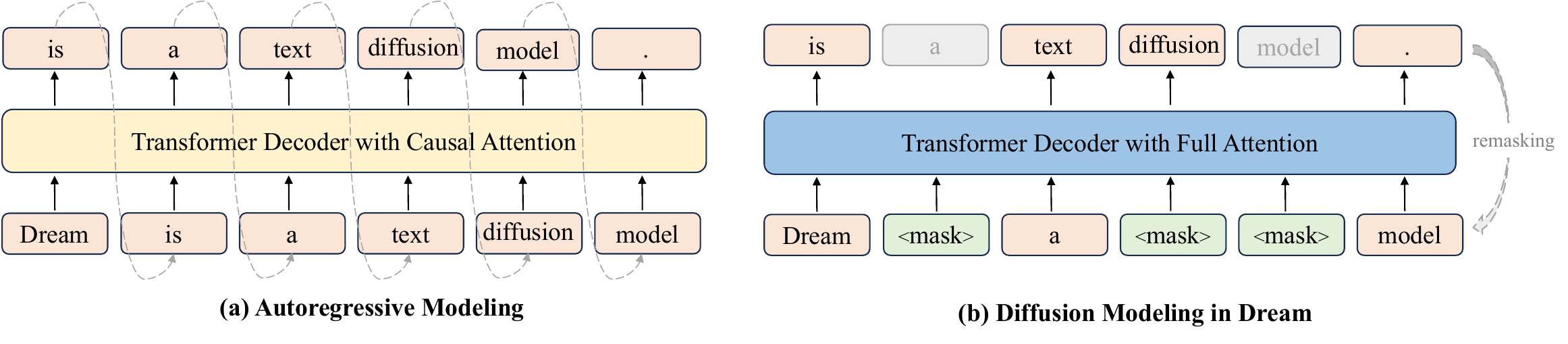}
    \vspace{-8mm}
    \caption{
        Comparison of autoregressive modeling and diffusion modeling in Dream. Dream predicts all the masked tokens in a shifted manner, allowing for maximumly architectural alignment and weight initialization with AR models.
    }
    \label{fig:framework}
    \end{figure}

\subsection{AR-based LLM Initialization}

Building on DiffuLLaMA~\citep{gong2025scalingdiffusionlanguagemodels}, we find that leveraging pre-trained weights from existing autoregressive (AR) models provides a non-trivial initialization strategy for diffusion language models. 
In \ourmodel, we adopt the "Shift Operation" strategy in ~\citet{gong2025scalingdiffusionlanguagemodels} to maintain the inherent shift operation characteristic of AR models during the transition to diffusion-based training.
Specifically, autoregressive models are trained so that the hidden representation at position $i$ encodes the information necessary to predict the token at position $i+1$. Rather than disrupting this learned positional relationship, our approach preserves this shift mechanism throughout the diffusion training process. The model continues to utilize hidden state $h_i$ to generate predictions for position $i+1$, which contrasts with conventional diffusion models that attempt to predict masked tokens at their original positions. This design choice ensures that the diffusion learning process remains aligned with the pretrained representations, preventing the transition from autoregressive to diffusion modeling from disrupting the foundational sequence understanding embedded in the pre-trained weights.

The AR initialization strategy offers computational advantages by building upon established sequence modeling capabilities rather than learning these representations from scratch, as evidenced by our experimental results in Section~\ref{sec:AR_init}. Furthermore, this approach maximizes the utility of existing pre-trained AR language models. Once superior AR LLMs become available, our framework enables rapid iteration toward improved diffusion models without requiring costly from-scratch training, effectively leveraging the collective efforts of the research community in autoregressive model development.

\begin{figure}[t]
    \centering
    \includegraphics[width=0.7\textwidth]{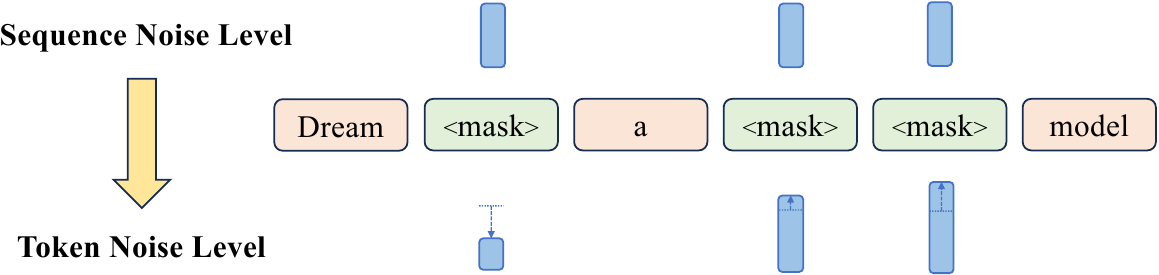}
    \caption{
        Illustration of the context-adaptive token-level noise rescheduling mechanism. Dream re-decides the noise level for each mask token by measuring its context informationness.
        }
    \label{fig:cart}
    \end{figure}
\subsection{Context-Adaptive Token-Level Noise Rescheduling}
\label{sec:cart}
Standard diffusion training for text applies uniform sentence-level timesteps, which ignores the varying contextual dependencies of individual tokens. In conventional discrete diffusion training, a single timestep $t$ is sampled to determine the overall noise level for the entire sequence, after which the model performs denoising on all the masked positions uniformly. For instance, when adopting the noise schedule $\alpha_t = 1-t$, the weighting term $w(t)$ in Equation~(\ref{eq:loss_dm}) becomes $\frac{1}{t}$. However, since learning occurs at the token level, the actual noise experienced by each token may deviate from the global timestep $t$. An example is shown in Figure~\ref{fig:cart}, a shared sentence-level $t$ is assigned to the three mask tokens in the standard discrete diffusion modeling, while the first token should inherently be associated with a lower noise level than subsequent tokens, as it benefits from richer contextual conditioning. This discrepancy leads to suboptimal learning, particularly for tokens with differing degrees of contextual information.

To address this limitation, we move from sequence-level noise scheduling to token-level noise scheduling. We introduce CART (\textbf{C}ontext-\textbf{A}daptive noise \textbf{R}escheduling at \textbf{T}oken-level), which measures the contextual informativeness for each token individually and dynamically reassigns appropriate noise levels based on the current context. This approach provides fine-grained, token-specific guidance that better reflects the varying difficulty and contextual dependencies across different positions in the sequence.
Specifically, the objective with CART is represented as follows:
\begin{equation}
    L(\theta) = -\E_{\rvx_0 \sim q(\rvx),t\sim \mathcal{U}(0,1), \rvx_t\sim q(\rvx_t|\rvx_0)} \sum_{n=1}^N \1_{[\mathrm{\rvx}_t^n=\texttt{MASK}]} w(t, \rvx_t, n) \log p_{\theta}(\rvx_0^n|\rvx_t),
\label{eq:loss_cart}
\end{equation}
where $w(t, \rvx_t, n)$ is a generalized term of $w(t)$ that can be flexibly designed. 
In practice, we employ a mixture of geometric distributions to quantify the information contribution of each clean token relative to the noised tokens,
which is defined as:
\begin{equation}
    w(t, \rvx_t, n)=\frac{1}{2}\sum_{i=1}^{N}\1_{[\mathrm{\rvx}_t^i\neq \texttt{MASK}]}\operatorname{Geo}(p,|n-i|-1),
\label{eq:cart}
\end{equation}
where $p \in (0,1]$ controls the sharpness of the geometric distribution. A smaller $p$ causes clean tokens to tend to contribute uniformly to all masked tokens in the sequence, while a larger $p$ forces clean tokens to have a greater influence on the nearby masked tokens.

\subsection{Training}
\paragraph{Model Architecture}
Dream models are built upon the Transformer architecture \citep{vaswani2017attention}, maintaining full compatibility with existing autoregressive language models while enabling diffusion-based training. Specifically, \ourmodel-7B adopts the same model configuration as Qwen2.5-7B~\citep{qwen2025qwen25technicalreport}.

\paragraph{Pretraining}
The training objective is to model the data distribution $p_\theta(\rvx_0)$. We estimate Equation~(\ref{eq:loss_dm}) via the Monte Carlo method for stochastic gradient descent training. Our training corpus comprises 580 billion tokens spanning text, mathematics, and code domains. The data is sourced from three primary repositories: Dolma v1.7~\citep{soldaini-etal-2024-dolma} for general domain text, OpenCoder~\citep{huang2025opencoderopencookbooktoptier} for programming and code datasets, and DCLM-Baseline~\citep{li2025dclm} for curated high-quality text and mathematical content. We exclusively utilize open-source datasets to demonstrate the effectiveness of our paradigm and ensure reproducibility.

\paragraph{Supervised Fine-Tuning} 
The SFT training objective is to model the conditional distribution $p_\theta(r_0|p_0)$, where $p_0$ is the prompt and $r_0$ denotes the response. During fine-tuning, noise is applied only to the response portion of the sequence, keeping the prompt intact to preserve instruction following capabilities. To align Dream with user instructions, we curate a dataset containing 1.8M instruction-response pairs from Tulu 3~\citep{lambert2025tulu3pushingfrontiers} and SmolLM 2~\citep{allal2025smollm2smolgoesbig}. 
\section{Experiments}
\subsection{Setup}

\paragraph{Evaluation Tasks}
We evaluated Dream on a comprehensive suite of benchmarks designed to assess its capabilities across multiple domains. For general language understanding, we employed tasks such as MMLU~\citep{hendrycks2020measuring}, BBH~\citep{suzgun2022challenging}, ARC-E~\citep{clark2018think}, ARC-C~\citep{clark2018think}, HellaSwag~\citep{zellers2019hellaswag}, WinoGrande~\citep{sakaguchi2021winogrande}, PIQA~\citep{bisk2020piqa}, and RACE~\citep{lai2017race} to test knowledge, reasoning, and commonsense abilities. For mathematical and scientific reasoning, we used GSM8K~\citep{cobbe2021training} and MATH~\citep{hendrycks2020measuring} for numerical computation and problem-solving, and GPQA~\citep{rein2023gpqa} for scientific understanding. Code generation abilities were evaluated through HumanEval~\citep{chen2021evaluating} and MBPP~\citep{austin2021program} to measure programming capabilities. The IFEval benchmark specifically measured the model's capacity to adhere to given instructions for instruction following assessment. 
For planning tasks, we evaluate performance on the Countdown and Sudoku tasks from \citet{ye2024beyond}, and Trip planning task from \citet{zheng2024naturalplanbenchmarkingllms}, which assess planning capabilities with specific objectives.

\begin{table}[t!]
    \centering
    \caption{ Comparison of base models on standard evaluation benchmarks. * indicates Dream 7B, LLaDA 8B, Qwen2.5 7B and LLaMA3 8B are evaluated under the same protocol. The best results are bolded and the second best are underlined. Our results are highlighted in \colorbox{lightblue}{blue}.
    }
    \label{tab:main_results}
    \scalebox{0.92}{
    \begin{tabular}{l c c c c | c c}
        \toprule
        Model & \cc{Dream 7B*} & LLaDA 8B* & Qwen2.5 7B* & LLAMA3 8B* & Mistral 7B & DeepSeek 7B \\
        \midrule
        Type &  \cc{}{Diffusion} & Diffusion & AR & AR & AR & AR \\
        Training Tokens & \cc{}0.6T & 2.3T & 18T &15T&- & 2T \\
        \midrule
        \multicolumn{7}{c}{\textbf{General Tasks}} \\
        \midrule
        MMLU & \cc{\underline{69.5 (5)}} & 65.9 (5) & \textbf{71.9 (5)} & 63.5 (5) & 60.1 (5) & 48.2 (5) \\
        BBH & \cc{\underline{57.9 (3)}} & 47.4 (3) & \textbf{63.9 (3)} & 62.7 (3) & - & 39.5 (3) \\
        ARC-E & \cc{}\textbf{83.9 (0)} & 71.8 (0) & 77.4 (0) & \underline{81.1 (0)} & 80.0 (0) & 67.9 (0) \\
        ARC-C & \cc{}\textbf{59.8 (0)} & 47.5 (0) & 51.5 (0) & \underline{53.6 (0)} & {55.5 (0)} & 48.1 (0) \\
        Hellaswag & \cc{\underline{73.3 (0)}} & 72.7 (0) & \textbf{79.0 (0)} & {78.9 (0)} & 81.3 (0) & 75.4 (0) \\ 
        WinoGrande & \cc{74.5 (5)} & 73.5 (5) & \underline{76.4 (5)} & \textbf{76.9 (5)} & 75.3 (0) & 70.5 (0) \\
        PIQA & \cc{75.8 (0)} & 74.8 (0) & \underline{79.8 (0)} & \textbf{81.3 (0)} & 83.0 (0) & 79.2 (0) \\ 
        RACE & \cc{}\textbf{44.7 (0)} & 38.7 (0) & \underline{41.9 (0)} & 39.2 (0) & - & {46.5 (5)} \\
        \midrule
        \multicolumn{7}{c}{\textbf{Mathematics \& Science}} \\
        \midrule
        GSM8K & \cc{\underline{77.2} (8)} & 70.9 (8) & \textbf{78.9 (8)} & 55.3 (8) & 52.1 (8) & 17.4 (8) \\
        MATH & \cc{\underline{39.6} (4)} & 30.7 (4) & \textbf{41.1 (4)} & 18.0 (4) & 13.1 (4) & 6.0 (4) \\
        GPQA & \cc{}\textbf{36.6 (5)} & 30.4 (5) & \underline{35.5 (5)} & 30.6 (5) & - & - \\
        \midrule
        \multicolumn{7}{c}{\textbf{Code}} \\
        \midrule
        HumanEval & \cc{}\textbf{57.9 (0)} & 32.9 (0) & \underline{56.7 (0)} & 35.4 (0) & 30.5 (0) & 26.2 (0) \\
        MBPP & \cc{\underline{56.2} (4)} & 39.0 (4) & \textbf{63.6 (4)} & 49.2 (4) & 47.5 (3) & 39.0 (3) \\
        \midrule
        \multicolumn{7}{c}{\textbf{Planning Tasks}} \\
        \midrule
        Countdown & \cc{}\textbf{16.0 (8)} & \underline{13.2 (8)} & 6.2 (8) & 3.7 (8) & - & - \\
        Sudoku & \cc{}\textbf{81.0 (8)} & \underline{46.0 (8)} & 21.0 (8) & 0.0 (8) & - & - \\
        Trip planning & \cc{}\textbf{17.8 (2)} & \underline{16.4 (2)} & 3.6 (2) & 8.7 (2) & - & - \\
        \bottomrule
    \end{tabular}
    }
    \end{table}

\subsection{Results of Dream-Base}
\paragraph{Superior planning performance of diffusion language models over autoregressive models} 
As shown in Table~\ref{tab:main_results}, Dream 7B maintains competitive general language capabilities while excelling in complex reasoning scenarios compared to Qwen2.5 7B, demonstrating the superior reasoning and planning abilities of diffusion models. For instance, on general tasks, Dream 7B achieves comparable performance with only modest gaps on MMLU and HellaSwag, while outperforming on reasoning tasks (ARC-E, ARC-C). The most striking advantages emerge in planning tasks, where Dream 7B substantially outperforms Qwen2.5 on Countdown (16.0 vs. 6.2), Sudoku (81.0 vs. 21.0), and Trip planning (17.8 vs. 3.6). Furthermore, we observe that other diffusion models (LLaDA) also consistently outperform Qwen2.5 on these tasks, providing additional evidence for the inherent advantages of diffusion modeling in planning capabilities.

\paragraph{Effectiveness of \ourmodel training approach} Compared to the previous state-of-the-art diffusion model (LLaDA 8B, which is trained from scratch), Dream 7B achieves remarkable performance improvements across all evaluation domains while using only 1/4 of the training data (0.6T vs 2.3T tokens). Dream 7B demonstrates significant gains on general tasks, with even more pronounced advantages in reasoning tasks. Notably, Dream 7B achieves substantial improvements on reasoning tasks (e.g., ARC-C improves from 47.5 to 59.8, GSM8K improves from 70.9 to 77.2). Additionally, Dream 7B excels in planning tasks with substantial margins over LLaDA: Countdown improves from 13.2 to 16.0, and Sudoku from 46.0 to 81.0. These results validate the effectiveness of our autoregressive model initialization strategy and context-adaptive noise rescheduling mechanism.
Notably, since Dream 7B is initialized from Qwen2.5 weights, we expect that with stronger autoregressive foundation models, our approach will enable the development of even more capable diffusion language models.

\subsection{Results of Dream-Instruct}
As an early exploration of diffusion LLM post-training, we perform lightweight supervised fine-tuning to align Dream with user instructions. Specifically, we curate 1.8M instruction-response pairs and fine-tune Dream for 3 epochs. The results of Dream-Instruct in Table~\ref{tab:sft_results} demonstrate DLLM's potential to match AR-based LLMs in instruction-following tasks, establishing a foundation for future advanced DLLM post-training recipes.

\begin{table}[h!]
\centering
\caption{Performance comparison after supervised fine-tuning on standard evaluation benchmarks.}
\label{tab:sft_results}
\vspace{1mm}
\begin{tabular}{lcccc}
\hline
{Model} & \cc{ }{{Dream 7B}} & {LLaDA 8B} & {{Qwen2.5 7B}} & {{LLaMA3 8B}} \\
\midrule
Type &\cc{ } {{Diffusion}} & {{Diffusion}} & {{AR}} & {{AR}} \\
Post-training & \cc{SFT} & SFT & SFT+RL & SFT+RL \\
Alignment pairs & \cc{1.8M} & 4.5M & 1M/0.15M & - \\
\midrule
MMLU & \cc{67.0} & 65.5 & 76.6 & 68.4 \\
MMLU-pro & \cc{43.3} & 37 & 56.3 & 41.9 \\
GSM8K & \cc{81.0} & 78.6 & 91.6 & 78.3 \\
Math & \cc{39.2} & 26.6 & 75.5 & 29.6 \\
GPQA & \cc{33.0} & 31.8 & 36.4 & 31.9 \\
HumanEval & \cc{55.5} & 47.6 & 84.8 & 59.8 \\
MBPP & \cc{58.8} & 34.2 & 79.2 & 57.6 \\
IFEval & \cc{62.5} & 59.9 & 74.7 & 49.7 \\
\hline
\end{tabular}
\end{table}

\subsection{Effect of AR Initialization}\label{sec:AR_init}
\begin{figure}[h]
    \centering
    \includegraphics[width=0.7\textwidth]{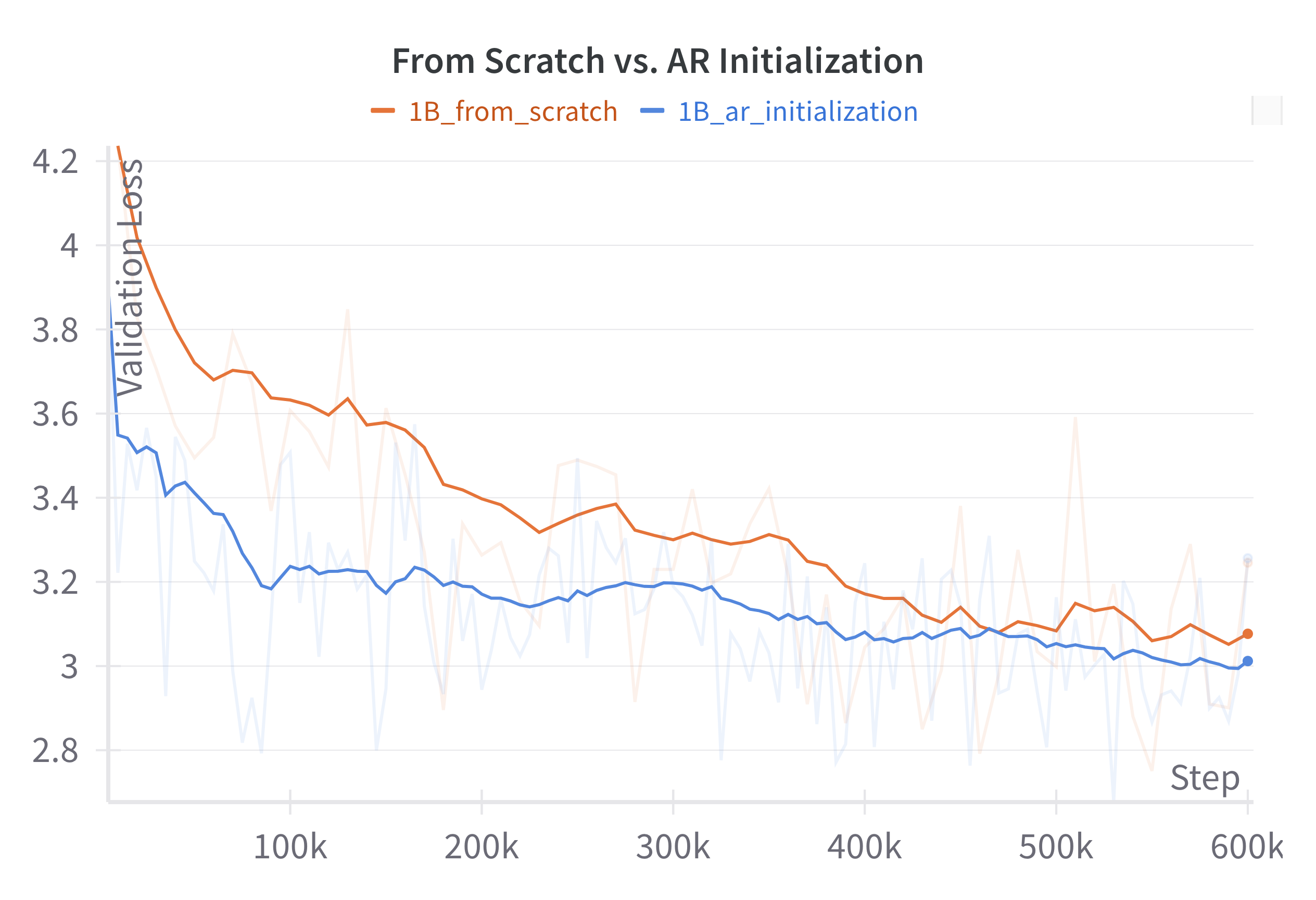}
    \caption{Loss comparison of the from-scratch and AR-initialization with LLaMA3.2 1B training on the Dream 1B models with 200B tokens. AR initialization also experiences a high loss at the beginning due to the transition from causal attention to full attention; however, this loss remains lower compared to training from scratch throughout the training.
    }
    \label{fig:scratch_vs_ar}
    \end{figure}

To validate the effectiveness of autoregressive (AR) initialization, we conducted preliminary experiments adapting Dream-1B from LLaMA3.2-1B. As demonstrated in Figure~\ref{fig:scratch_vs_ar}, the AR initialization strategy proves particularly effective during the early stages of training, providing significant advantages over training from scratch. This observation shows that the existing left-to-right knowledge in AR models provides a strong foundation that accelerates the diffusion model's development of any-order generation capabilities. This acceleration significantly reduces both the number of tokens and computational resources required for pretraining, making our approach highly efficient compared to training diffusion language models from scratch. 

Besides, our analysis reveals that the learning rate plays a critical role in preserving the beneficial properties inherited from AR initialization. An excessively high learning rate can rapidly degrade the left-to-right linguistic knowledge embedded in the initial weights, thereby diminishing the advantages of AR initialization. Conversely, an overly conservative learning rate may impede the model's ability to effectively learn the diffusion process. Through systematic hyperparameter tuning, we carefully calibrated the learning rate alongside other training parameters to achieve optimal performance.

\subsection{Discussion}
\subsubsection{Analysis of Planning Ability}
\begin{figure}[ht]
    \centering
    \includegraphics[width=\textwidth]{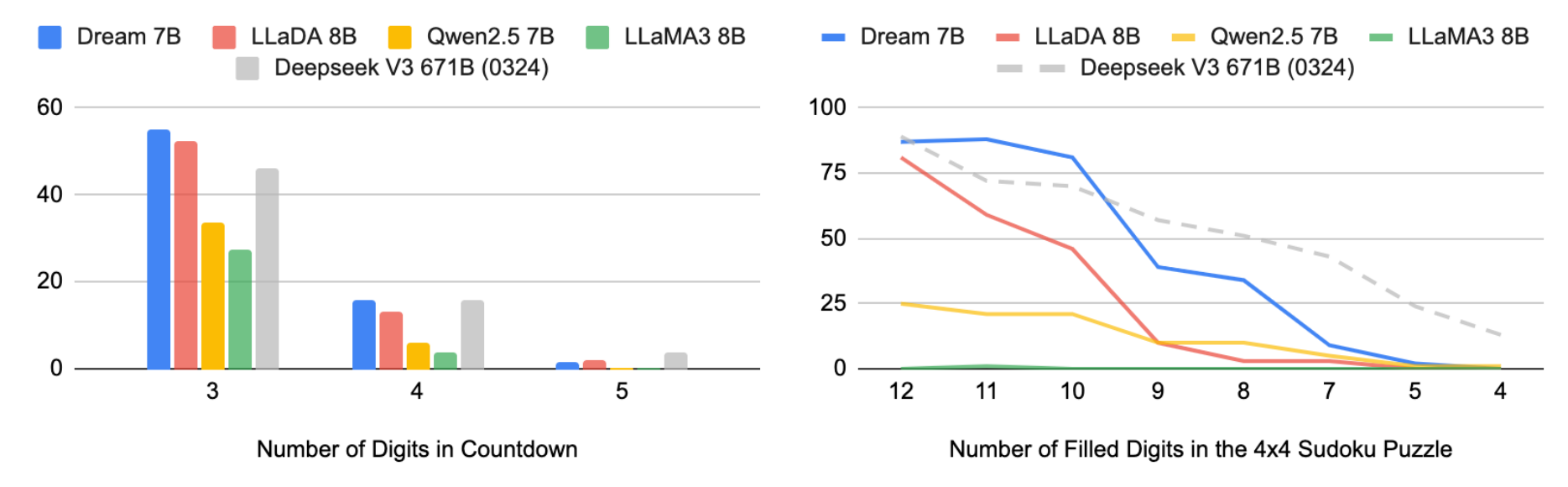}
    \caption{Results on Countdown and Sudoku when varying planning difficulty.
    }
    \label{fig:planning}
    \end{figure}
    
~\citealt{ye2024beyond,ye2025implicit} demonstrated that text diffusion models exhibit superior planning capabilities in small-scale, task-specific contexts. However, it remained unclear whether these advantages extend to general-purpose, scaled diffusion models. In this work, we scale up diffusion language models and systematically evaluate their constraint-based planning abilities using Dream 7B.
We evaluated Dream on the Countdown and Sudoku tasks from~\cite{ye2024beyond}. Our experimental comparison included Dream 7B alongside LLaDA 8B, Qwen2.5 7B, and LLaMA3 8B, with the DeepSeek V3-671B (0324) serving as a reference point. All models were evaluated in a few-shot setting without task-specific training.

\begin{figure}[h]
    \centering
    \includegraphics[width=1\textwidth]{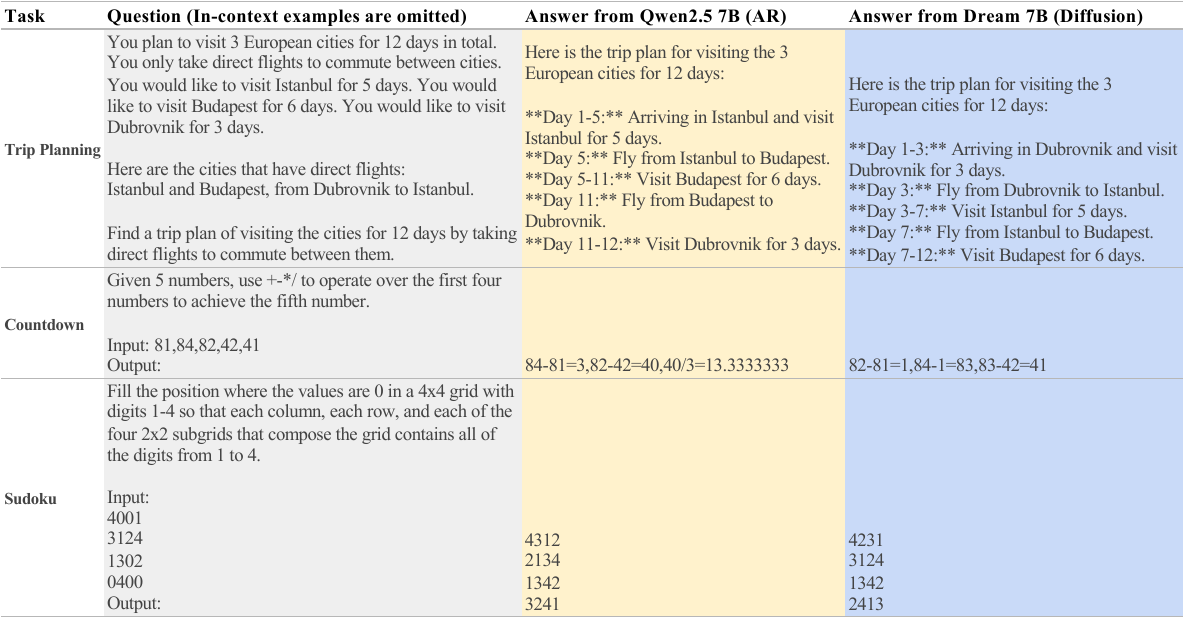}
    \caption{Examples generated by Qwen2.5 7B and Dream 7B.
    }
    \label{fig:demo_planning}
    \end{figure}

The results shown in Figure~\ref{fig:planning} demonstrate that Dream consistently outperforms other similarly-sized baseline models across both tasks. Notably, Dream 7B significantly exceeds the performance of comparably-sized autoregressive models (LLaMA3 8B and Qwen2.5 7B) and, in certain cases (e.g., Countdown3), even surpasses the much larger DeepSeek-V3-671B (0324), despite the latter having orders of magnitude more parameters. This substantial performance advantage suggests that diffusion language models are inherently more effective at solving problems involving multiple constraints or specific objective optimization. For detailed qualitative examples illustrating the performance differences between Qwen2.5 7B and Dream 7B across various planning scenarios, please refer to Figure~\ref{fig:demo_planning}.

\subsubsection{Quality-Speed Trade-Offs}
Diffusion language models provide a unique advantage through their adjustable inference process, where the number of diffusion timesteps can be modified to balance computational speed and output quality. Reducing the number of timesteps speeds up inference but may reduce generation quality, while increasing timesteps improves output quality at the cost of additional computation.

\begin{figure}[h]
    \centering
    \includegraphics[width=0.7\textwidth]{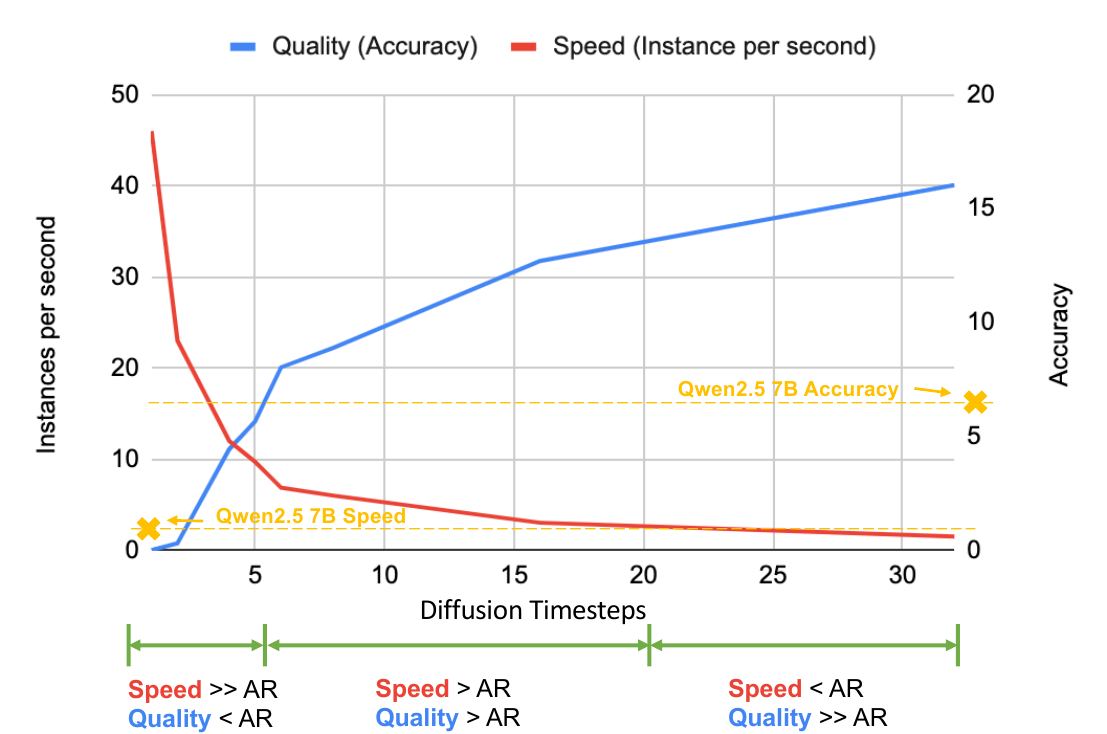}
    \caption{Quality-speed comparison on the Countdown task for Dream 7B and Qwen2.5 7B. By adjusting the diffusion timesteps, the performance of Dream can be flexibly tuned for either speed or quality.
    }
    \label{fig:quality_speed}
    \end{figure}

To demonstrate this flexibility, we conduct a systematic quality-speed analysis on the Countdown task, comparing Dream 7B against Qwen2.5 7B across varying timestep configurations, as illustrated in Figure~\ref{fig:quality_speed}. The experimental results demonstrate that Dream's performance exhibits a controllable trade-off between computational efficiency and generation quality through timestep adjustment, providing users with fine-grained control over the inference process. Notably, when diffusion steps are set between 5-20, \ourmodel~achieves superior performance in both speed and quality compared to Qwen2.5 7B.

This timestep-based approach introduces an additional dimension for inference-time scaling \citep{snell2024scaling,muennighoff2025s1,geiping2025scaling}, which works alongside existing techniques such as chain-of-thought reasoning used in large language models like OpenAI o1~\citep{jaech2024o1} and DeepSeek R1~\citep{guo2025deepseekR1}. This adjustable computation-quality trade-off represents a key advantage that sets diffusion models apart from traditional autoregressive models.

\subsubsection{Arbitrary Order Generation}
Diffusion models exhibit remarkable inference flexibility, setting them apart from traditional autoregressive (AR) models that typically generate sequences in a fixed left-to-right order. This flexibility stems from their inherent capability to synthesize outputs in arbitrary orders, opening up a wider range of diverse user queries and interaction patterns.

Dream 7B supports completion and infilling tasks naturally, without requiring specialized training. Key examples of this flexibility include:
\begin{itemize}
    \item \textbf{Completion:} Users can provide an initial piece of text, and the model can seamlessly continue or complete it.
    \item \textbf{Infilling:} Diffusion models can fill in missing segments within a given text. This can be unconstrained or guided by specific requirements, such as ensuring the generated infill leads to an exact ending sentence.
    \item \textbf{Configurable decoding order:} Beyond these task-specific flexibilities, users can also exert control over the decoding behavior itself. By adjusting decoding hyperparameters, the generation process can be tailored to mimic the more structured, left-to-right generation characteristic of AR models, or conversely, it can introduce varying degrees of randomness in the decoding order, moving towards a partially or even fully random-order synthesis.
\end{itemize}

This adaptability in how the output is constructed, not just what is generated, allows diffusion models to cater to different preferences and task requirements, making them versatile tools for a broad spectrum of generation scenarios.
We provide dynamic demonstrations of these capabilities at \url{https://hkunlp.github.io/blog/2025/dream/}.

\section{Conclusion}
Dream 7B represents a significant step forward in diffusion-based language modeling, achieving competitive performance with state-of-the-art autoregressive models while providing unique capabilities for flexible text generation. Our experiments reveal that diffusion language models excel particularly in constraint-satisfaction and planning tasks, where their bidirectional nature and iterative refinement process offer clear advantages over traditional left-to-right generation approaches.

The model's ability to perform arbitrary-order generation enables novel applications in completion and infilling tasks, while the adjustable timestep configurations provide users with fine-grained control over the quality-speed trade-off during inference. These capabilities, combined with the demonstrated effectiveness of initializing from pre-trained autoregressive models, establish a practical pathway for developing more flexible and capable language models.
Future work will explore advanced post-training recipes, longer context capabilities, and applications to specialized domains requiring enhanced planning and reasoning abilities. 
\section*{Acknowledgments}
We thank Chengwu Cai for his support and helpful discussion. We also acknowledge the open-source community for providing high-quality datasets and evaluation frameworks. 
This research was supported in part by the joint research scheme of the National Natural Science Foundation of China (NSFC) and the Research Grants Council (RGC) under grant number
N\_HKU714/21.

\bibliographystyle{plainnat}
\bibliography{custom}

\end{document}